\pdfoutput=1

\documentclass[11pt]{article}

\usepackage{acl2023}

\usepackage{times}
\usepackage{latexsym}

\usepackage[T1]{fontenc}

\usepackage[utf8]{inputenc}

\usepackage{microtype}

\usepackage{inconsolata}

\usepackage{tikz}
\usetikzlibrary{positioning}
\usetikzlibrary{bayesnet}
\usetikzlibrary{arrows}
\usetikzlibrary{backgrounds}
\usepackage{color}
\usepackage{graphicx}
\usepackage{caption}
\usepackage{subcaption}

\usepackage{amsmath}
\usepackage{amssymb}

\usepackage{cleveref}

\usepackage{todonotes}
\usepackage{hhline}
\usepackage{colortbl}

\usepackage{tabularx}

\usepackage{relsize}
\usepackage{multirow}

\DeclareMathOperator*{\argmax}{arg\,max}

\DeclareMathOperator*{\tf}{tf}
\DeclareMathOperator*{\idf}{idf}

\def\fs{\kern 0.5em}

\newcommand\Tstrut{\rule{0pt}{2.6ex}}         
\newcommand\Bstrut{\rule[-1.6ex]{0pt}{0pt}}   

\usepackage{mathtools}

\DeclarePairedDelimiterX{\infdivx}[2]{\big[}{\big ]}{%
  #1\;\delimsize\|\;#2%
}
\newcommand{\infdiv}{D_{KL}\infdivx}

\title{Attributable and Scalable Opinion Summarization}

\author{{Tom Hosking \qquad Hao Tang \qquad Mirella Lapata} \\
  Institute for Language, Cognition and Computation \\
  School of Informatics, University of Edinburgh \\
  10 Crichton Street, Edinburgh EH8 9AB\\
  \texttt{tom.hosking@ed.ac.uk}\quad \texttt{hao.tang@ed.ac.uk} \quad \texttt{mlap@inf.ed.ac.uk}}

\begin{document}
\maketitle
\begin{abstract}
We propose a method for unsupervised opinion summarization that encodes sentences from customer reviews into a hierarchical discrete latent space, then identifies common opinions based on the frequency of their encodings. We are able to generate both abstractive summaries by decoding these frequent encodings, and extractive summaries by selecting the sentences assigned to the same frequent encodings. Our method is attributable, because the model identifies sentences used to generate the summary as part of the summarization process. It scales easily to many hundreds of input reviews, because aggregation is performed in the latent space rather than over long sequences of tokens. We also demonstrate that our appraoch enables a degree of control, generating aspect-specific summaries by restricting the model to parts of the encoding space that correspond to desired aspects (e.g., location or food). Automatic and human evaluation on two datasets from different domains demonstrates that our method generates summaries that are more informative than prior work and better grounded in the input reviews.
\end{abstract}

\section{Introduction}

Online review websites are a useful resource when choosing which hotel to visit or which product to buy, but it is impractical for a user to read hundreds of reviews. There has been significant interest in methods for automatically generating summaries or {meta-reviews} that aggregate the diverse opinions contained in a set of customer reviews about an \textit{entity} (e.g., a product, hotel or restaurant) into a single summary. 

Early work on opinion summarization extracted reviewers' sentiment about specific features \cite{10.1145/1014052.1014073} or selected salient sentences from reviews based on centrality \cite{lexrank}, while more recent methods based on neural models have used sentence selection in learned feature spaces \cite{angelidis-etal-2021-extractive,basu-roy-chowdhury-etal-2022-unsupervised} or abstractive summarizers that generate novel output \cite{brazinskas-etal-2020-unsupervised,brazinskas-etal-2021-learning,amplayo-etal-2021-aspect, amplayo2021unsupervised,iso-etal-2021-convex-aggregation,coavoux-etal-2019-unsupervised}.

Following \citet{ganesan-etal-2010-opinosis}, we define opinion summarization, or \textit{review aggregation}, as the task of generating a textual summary that reflects frequent or popular opinions expressed in a large number of reviews about an entity. Systems are \textit{extractive} if they select sentences or spans from the input reviews to use as the summary, or \textit{abstractive} if they generate novel output. Review aggregation is challenging for a number of reasons. Firstly, it is difficult to acquire or create reference summaries, so models are almost always trained without access to gold standard references \cite[][\textit{inter alia}.]{angelidis-etal-2021-extractive,amplayo2021unsupervised}. Secondly, popular entities may have hundreds of reviews, which can cause computational difficulties if the approach is not \textit{scalable}. Finally, good summaries should be \textit{abstractive} and not contain unnecessary detail, but should also not hallucinate false information. Ideally, a summarization system should be \textit{attributable}, offering some evidence to justify its output. 

\begin{figure}[t!]
    \centering
    \includegraphics[width=0.48\textwidth]{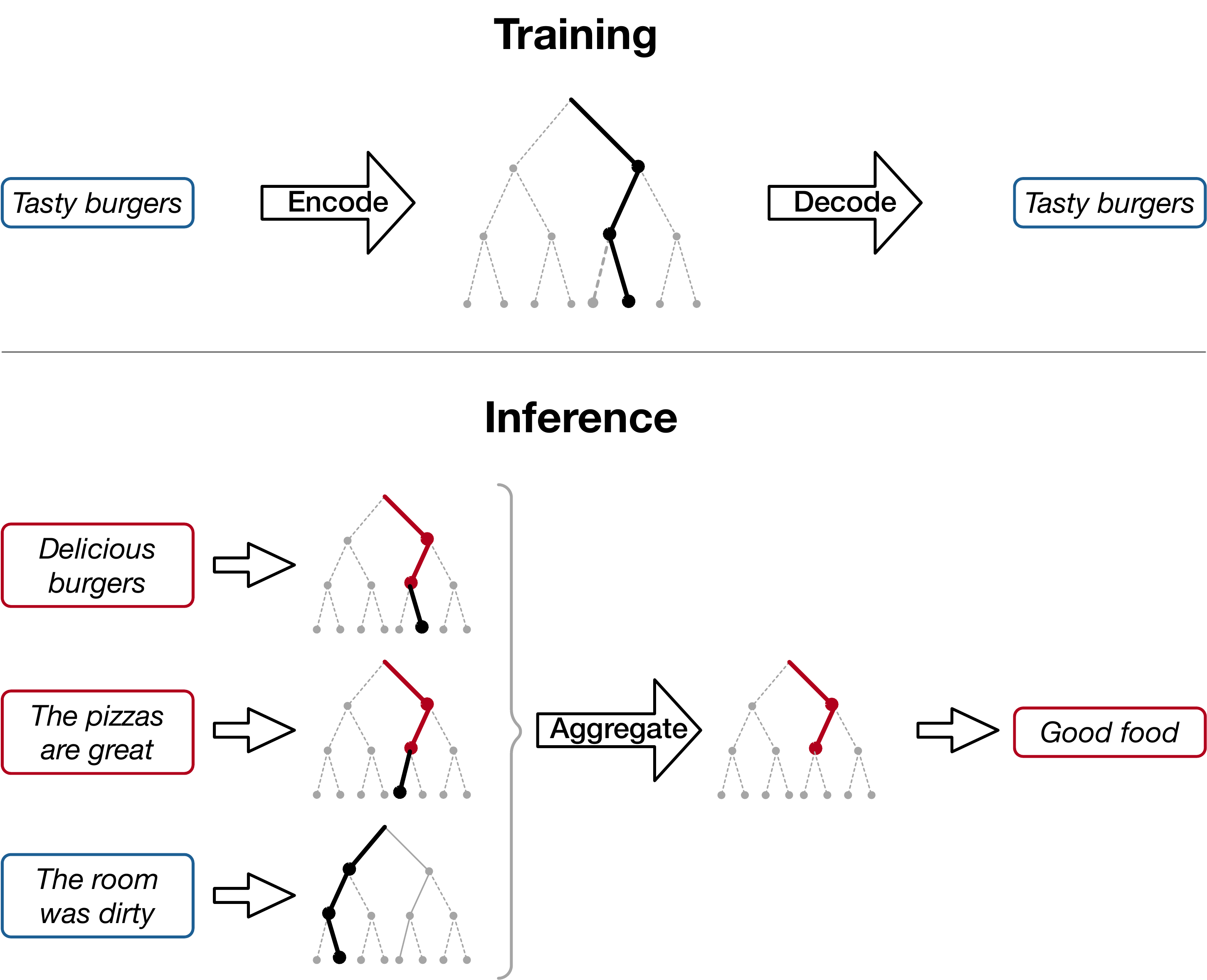}  
    \vspace{-0.2cm}
    \caption{\textsc{Hercules} is trained to encode sentences from reviews as paths through a hierarchical discrete latent space (top). At inference time, we encode all sentences from the input reviews, and identify frequent paths or subpaths to use for the summary (bottom). The consensus opinion from the three example inputs is that the food is good, so the subpath shown in red is repeated; decoding it should result in an output like "Good food". }
    \vspace{-0.2cm}
    \label{fig:idealised}
    \vspace{-0cm}
\end{figure}

Previous work has either been exclusively extractive (which is inherently attributable and often scalable but leads to unnecessarily specific summaries) or exclusively abstractive \cite[which often scales poorly and hallucinates, e.g.,][]{brazinskas-etal-2020-unsupervised} . We propose a hybrid method, that produces abstractive summaries accompanied by references to input sentences which act as evidence for each output sentence, allowing us to verify which parts of the input reviews were used to produce the output. Depicted in \Cref{fig:idealised}, we first learn to encode natural language sentences from reviews as paths through a hierarchical discrete latent space. Then, given multiple review sentences about a specific entity, we identify common subpaths that are shared among many inputs, and decode them back to natural language, yielding the output summary. The sentences whose encodings contain the selected subpaths (shown in red in \Cref{fig:idealised}) act as evidence for that generated sentence.

Our approach, \textsc{Hercules}, is unsupervised and does not need reference summaries during training, instead relying on properties of the encoding space induced by the model. Since the aggregation process occurs in encoding space rather than over long sequences of tokens, \textsc{Hercules} is highly scalable. Generated summaries are accompanied by supporting evidence from input reviews, making \textsc{Hercules} attributable. It also offers a degree of controllability: we can generate summaries that focus on a specific aspect of an entity (e.g., location) or sentiment by restricting aggregation to subpaths that correlate with the desired property.

Our contributions are as follows:
\begin{itemize}
    \item We propose a method for representing natural language sentences as paths through a hierarchical discrete latent space (\Cref{sec:hrqvae}).
    \item We exploit the properties of the learned hierarchy to identify common opinions from input reviews, and generate abstractive summaries alongside extractive \emph{evidence sets} (\Cref{sec:opagg}).
    \item We conduct extensive experiments on two English datasets covering different domains, and show that our method outperforms previous state-of-the-art approaches, while offering the additional advantages of attributability and scalability (Sections \ref{sec:experiments} and \ref{sec:results}).
\end{itemize}

\section{Hierarchical Quantized Autoencoders}
\label{sec:hrqvae}

A good review aggregation system should identify frequent or common opinions, while abstracting away the details unique to a specific review. This joint requirement motivates our choice of a hierarchical discrete encoding: the discretization allows us to easily identify repeated opinions by counting them, while the hierarchy allows the model to encode high-level information (aspect, sentiment etc.) separately to specific details and phrasings.


\subsection{Probabilistic Model}

Let $\textbf{y}$ be a sentence, represented as a sequence of tokens. We assume that the semantic content of $\textbf{y}$ may be encoded as a set of discrete latent variables or \textit{codes} $q_{1:D} \in [1, K]$. Further, we assume that the $q_{1:D}$ are ordered \textit{hierarchically}, such that $q_1$ represents high level information about the sentence (e.g., the aspect or overall sentiment) whereas $q_D$ represents fine-grained information (e.g., the specific phrasing or choice of words used). The codes $q_{1:D}$ can be viewed as a single \textit{path} through a hierarchy or tree as depicted in \Cref{fig:idealised}, where each intermediate and leaf node in the tree corresponds to a sentence $\textbf{y}$.

Thus, our generative model factorises as
\begin{multline}
\vspace{-.2cm}
    p(\textbf{y})
    = \sum_{q_{1:D}}p(\textbf{y}| q_{1:D}) \times \prod
\limits_{d=1}^D  p(q_d)
\end{multline}
and the posterior factorises as
\vspace{-.2cm}
\begin{multline}
    \phi(q_{1:D} | \textbf{y}) =
     \phi(q_1 | \textbf{y}) \times \prod \limits_{d=2}^D \phi(q_d | q_{< d}, \textbf{y}).
\end{multline}

The training objective is given by
\begin{multline} \label{eq:finalobjective}
    \text{ELBO} = \mathbb{E}_{\phi}\big [-\log p(\textbf{y} |  q_{1:D}) \big ] \\
      + \beta_{KL} \sum\limits_{d=1}^D \infdiv{\phi(q_d | \textbf{y}) }{ p(q_d) }
\end{multline}
where $q_d \sim \phi(q_d|\textbf{y})$ and $\beta_{KL}$ determines the weight of the KL term. We choose a uniform prior for $p(q_d)$.


\subsection{Neural Parameterization}

The latent codes $q_{1:D}$ are discrete, but most neural methods operate in continuous space. We therefore need to define a mapping from the output $\textbf{z} \in \mathbb{R}^{\mathbb{D}}$ of an encoder network $\phi(\textbf{z} | \textbf{y})$ to $q_{1:D}$, and vice versa for a decoder $p(\textbf{y} | \textbf{z})$. Similiar to Vector Quantization \cite[VQ,][]{vqvae}, we learn a codebook $\textbf{C}_d \in \mathbb{R}^{K \times \mathbb{D}}$, which maps each discrete code to a continuous embedding $\textbf{C}_d(q_d) \in \mathbb{R}^{\mathbb{D}}$.

 Similar to HRQ-VAE \cite{hosking-etal-2022-hierarchical}, since the $q_{1:D}$ are intended to represent hierarchical information, the distribution over codes at each level is a softmax distribution with scores~$s_d$ given by the L2 distance from each of the codebook embeddings to the residual error between the input and the cumulative embedding from all previous levels, 
\begin{align}
\vspace{-.2cm}
\hspace{-0.25cm} s_d(q) = - \left( \left[\textbf{x} - \sum\limits_{d'=1}^{d-1} \textbf{C}_{d'}(q_{d'}) \right ] - \textbf{C}_d(q)  \right ) ^2\hspace{-0.25cm} .
\vspace{-.2cm}
\end{align}
During inference, we set $q_d = \argmax (s_d)$.

Given a path $q_{1:D}$, the input to the decoder $\textbf{z}$ is given by the inverse of the decomposition process,
\begin{align}
\vspace{-.2cm}
\textbf{z} = \sum\limits_{d=1}^{D} \textbf{C}_d(q_{d}).
\vspace{-.2cm}
\end{align}
The embeddings at each level can be viewed as refinements of the (cumulative) embedding so far, or alternatively as selecting the centroid of a subcluster within the current cluster. Importantly, it is not necessary to specify a path to the complete depth $D$; a \textit{subpath} $q_{1:d}$ ($d<D$) still results in a valid embedding $\textbf{z}$. We can therefore control the specificity of an encoding by varying its depth.



\subsection{Training Setup} 

We use the Gumbel reparameterization
\cite{jang2016categorical,maddison2017concrete,sonderby2017continuous}
to sample from the distribution over $q_{1:D}$. To encourage the model to explore the full codebook, we decay the Gumbel temperature~$\tau$ according to the schedule given in \Cref{app:replication}. We approximate the expectation in
\Cref{eq:finalobjective} by sampling from the training set and
updating via backpropagation \cite{kingma2013autoencoding}.


\paragraph{Initialization Decay and Norm Loss} Smaller perturbations in encoding space should result in more fine-grained changes in the information they encode. Therefore, we encourage \textit{ordering} between the levels of hierarchy (such that lower levels encode more fine-grained information) by initialising the codebook with a decaying magnitude, such that deeper embeddings have a smaller norm than those higher in the hierarchy. Specifically, the norm of the embeddings at level $d$ is weighted by a factor~$(\alpha_{init}) ^ {d-1}$. We also include an additional loss $\mathcal{L}_{NL}$ to encourage deeper embeddings to remain fine-grained during training, 
\begin{multline*}
\vspace{-.4cm}
\hspace{-.35cm}
    \mathcal{L}_{NL} = \frac{\beta_{NL}}{D} \sum \limits_{d=2}^{D} \big [ \max \big ( \gamma_{NL} \frac{ \lvert\lvert \textbf{C}_d \rvert\rvert_2 }{ \lvert\lvert \textbf{C}_{d-1} \rvert\rvert_2 } , 1\big ) - 1 \big ]^2,
\vspace{-.4cm}
\end{multline*}
where $\gamma_{NL}$ determines the relative scale between levels and $\beta_{NL}$ controls the strength of the loss.

\paragraph{Depth Dropout} To encourage the hierarchy within the encoding space to correspond to hierarchical properties of the output, we truncate at each level during training with some probability $p_{depth}$ \cite{hosking-etal-2022-hierarchical,10.1109/TASLP.2021.3129994}. The output of the quantizer is then given by 
\begin{align}
\vspace{-.4cm}
\textbf{z}_{syn} = \sum \limits_{d=1}^D \left ( \textbf{C}_d(q_d) \prod\limits_{d'=1}^d \gamma_{d'} \right ) ,
\vspace{-.4cm}
\end{align}
where $\gamma_h \sim \text{Bernoulli}(1 - p_{depth})$. This means that the model is sometimes trained to reconstruct the output based only on a \textit{partial} encoding of the input, and should learn to cluster similar outputs together at each level in the hierarchy.

\paragraph{Denoising Objective} To encourage the model to group sentences according to their meaning rather than their syntactic structure, we use a denoising objective as a form of weak supervision. The model is trained to generate a target sentence from a different source sentence that has similar meaning but different surface form. For example, given the target sentence ``We chose this hotel for price/location.'',   a source might be ``I chose this hotel for its price and location.''. The source sentences are retrieved automatically from other reviews in the training data using tf-idf \cite{tfidf} over bigrams; we select the top 5 most similar sentences for each target sentence with a minimum similarity of 0.6, and restrict to retrieving from reviews that have ratings equal to the target. 


\section{Aggregating Reviews in Encoding Space}
\label{sec:opagg}
\vspace{-.1cm}
So far, we have described a method for mapping from a sentence $\textbf{y}$ to a path $q_{1:D}$ and vice versa. We can now exploit the hierarchical property of the latent space to generate summaries.

Recall that the goal of review aggregation is to identify the majority or frequent opinions from a set of diverse inputs. This corresponds to identifying paths (or subpaths) in encoding space that are shared among many inputs. A simplified version of this process is depicted in the lower block of \Cref{fig:idealised}; each sentence $\textbf{y}^{(i)}$ in the input reviews is mapped to a path $q_{1:D}^{(i)}$ through the latent space. Summarizing these sentences is then reduced to the task of selecting a set of common subpaths, e.g., the subpath highlighted in red in \Cref{fig:idealised}, which is shared between two out of three inputs.

\paragraph{Subpath Selection}
\label{sec:summ_construction}

A simple approach would be to select the most frequent subpaths, but this would almost always result in high-level paths with $d=1$ being selected (since every occurrence of a path $q_{1:d}$ entails an occurrence of all subpaths $q_{1:d'}, d'<d$). In practice there is a tradeoff between frequency and specificity. Additionally, good summaries often exhibit structure; they generally include high-level comments, alongside more specific comments about details that particularly differentiate the current entity from others. Indeed, some datasets \cite[e.g., AmaSum,][\Cref{sec:datasets}]{brazinskas-etal-2021-learning} were constructed by scraping overall `verdicts' and specific `pros and cons' from review websites. We therefore reflect this structure and propose both a `generic' and `specific' method for selecting subpaths.

To select \textit{generic} subpaths, we construct a probability tree from the set of input sentence encodings, with the node weights set to the observed path frequency $p(q_{1:d})$. Then, we iteratively prune the tree, removing the lowest probability leaves until all leaf weights exceed a threshold, $\min\big(p(q_{1:d})\big) > 0.01$. Finally, we select the leaves with the top~$k$ weights to use for the summary. Empirically, this approach often selects paths with depth $d=1$, but allows additional flexibility when a deeper subpath is particularly strongly represented.

Similar to \citet{iso-etal-2022-comparative} we argue that the \textit{specific} parts of the summary should also be comparative, highlighting details that are unique to the current entity. Thus, tf-idf \cite{tfidf} is a natural choice; we treat each path (and all its parent subpaths) as terms. We assign scores to each subpath $q_{1:d}$ proportional to its frequency within the current entity, and inversely proportional to the number of entities in which the subpath appears,
\begin{multline}
\vspace{-.4cm}
    \text{score}(q_{1:d}) = \tf(q_{1:d}) \times \log\big(\idf(q_{1:d})\big).
    \vspace{-.4cm}
    \label{eq:scoring}
\end{multline}
Again, we select the subpaths with the top~$k$ scores to use for the summary.

The overall summary is the combination of the selected generic and specific subpaths. The abstractive natural language output is generated by passing the selected subpaths as inputs to the decoder.

\paragraph{Attribution} Each sentence in the generated summary has an associated subpath. By identifying all inputs which share that subpath, we can construct an \textit{evidence set} of sentences that act as an explanation or justification for the generated output.

\paragraph{Scalability} Since the aggregation is performed in encoding space, our method scales linearly with the number of input sentences (compared to quadratic scaling for Transformer methods that take a long sequence of all review sentences as input, e.g., \citet{instructgpt}), and can therefore handle large numbers of input reviews. In fact, since we identify important opinions using a frequency-based method, our system does not perform well when the number of input reviews is small, since there is no strong signal as to which opinions are common.

\paragraph{Controlling the Output} Given an aspect $a$ (e.g.,~`service') we source a set of keywords $\mathbb{K}_a$ (e.g.,~`staff, friendly, unhelpful, concierge') associated with that aspect \cite{angelidis-etal-2021-extractive}. We label each sentence in the training data with aspect $a$ if it contains any of the associated keywords $\mathbb{K}_a$, then calculate the probability distribution over aspects for each encoding path, $p(a|q_{1:D})$. We can modify the scoring function in \Cref{eq:scoring}, multiplying the subpath scores during aggregation by the corresponding likelihood of a desired aspect, thereby upweighting paths relevant to that aspect,
\begin{multline}
\vspace{-.4cm}
    \text{score}_{asp}(q_{1:d}) = \tf(q_{1:d}) \times \log\big(\idf(q_{1:d})\big) \\ \times p(a|q_{1:D}).
    \vspace{-.4cm}
\end{multline}

We can also control for the sentiment of the summary; for the case where reviews are accompanied by ratings, we can label each review sentence (and its subpath) with the rating $r$ of the overall review, and reweight the subpath scores during aggregation by the likelihood of the desired rating $p(r|q_{1:D})$.



\section{Experimental Setup}
\label{sec:experiments}

\subsection{Datasets}
\label{sec:datasets}

We perform experiments on two datasets from two different domains. \textsc{\textbf{Space}} \cite{angelidis-etal-2021-extractive} consists of hotel reviews from TripAdvisor, with 100 reviews per entity. It includes reference summaries constructed by human annotators, with multiple references for each entity. It also includes reference \textit{aspect-specific} summaries, which we use to evaluate the controllability of \textsc{Hercules}.

\textbf{AmaSum} \cite{brazinskas-etal-2021-learning} consists of reviews of Amazon products from a wide range of categories, with an average of 326 reviews per entity. The reference summaries were collected from professional review websites, and therefore are \emph{not grounded} in the input reviews. The references in the original dataset are split into `verdict', `pros' and `cons'; we construct single summaries by concatenating these three. We filter the original dataset down to four common categories (Electronics, Shoes, Sports \& Outdoors, Home \& Kitchen), and evaluate on a subset of 50 entities, training separate models for each. All systems were trained and evaluated on the same subsets.

\subsection{Comparison Systems}

We compare with a range of baseline and comparison systems, both abstractive and extractive. For comparison, we construct extractive summaries using \textsc{Hercules} by selecting the centroid from each evidence set based on ROUGE-2 F1 score.


We select a \textbf{random review} from the inputs as a lower bound. We also select the \textbf{centroid} of the set of reviews, according to ROUGE-2 F1 score. We include an extractive \textbf{oracle} as an upper bound, by selecting the input sentence with highest ROUGE-2 similarity to each reference sentence.

\textbf{Lexrank} \cite{lexrank} is an unsupervised extractive method using graph-based centrality scoring of sentences.

\textbf{QT} \cite{angelidis-etal-2021-extractive} uses vector quantization to map sentences to a discrete encoding space, then generates extractive summaries by selecting representative sentences from clusters.

\textbf{SemAE} \cite{basu-roy-chowdhury-etal-2022-unsupervised} is an extractive method that extends QT, relaxing the discretization and encoding sentences as mixtures of learned embeddings.

\textbf{CopyCat} \cite{brazinskas-etal-2020-unsupervised} is an abstractive approach that models sentences as observations of latent variables representing entity opinions.

\textbf{InstructGPT} \cite{instructgpt} is a Large Language Model that generates abstractive summaries via prompting. We use the variant `text-davinci-002'; training details are not public, but it is likely that it was tuned on summarization tasks, and potentially had access to the evaluation data for both \textsc{Space} and AmaSum during training.

\textbf{BiMeanVAE} and \textbf{COOP} \cite{iso-etal-2021-convex-aggregation} are abstractive methods that encode full reviews as continuous latent vectors, and take the average (BiMeanVAE) or an optimised combination (COOP) of review encodings.

Finally, for aspect specific summarization we compare to \textbf{AceSum} \cite{amplayo-etal-2021-aspect}. AceSum uses multi-instance learning to induce a synthetic dataset of review/summary pairs with associated aspect labels, which is then used to train an abstractive summarization model.

Most of the abstractive methods are not scalable and have upper limits on the number of input reviews. CopyCat and InstructGPT have a maximum input sequence length, while COOP exhaustively searches over combinations of input reviews. We use 8 randomly selected reviews as input to CopyCat and COOP, and 16 for InstructGPT.

\newcolumntype{H}{>{\setbox0=\hbox\bgroup}c<{\egroup}@{}}

\begin{table*}[ht!]
    \centering
\small
    \begin{tabular}{@{~}cl||Hrr|rrr||Hrr|rrr}
     & & \multicolumn{6}{c||}{\textbf{\textsc{Space}}} & \multicolumn{6}{c}{\textbf{{AmaSum} (4 domains)}} \\
    & \textbf{System} & {R-1} $\uparrow$ & {R-2} $\uparrow$  & {{R-L}} $\uparrow$  & {QA} $\uparrow$ & {SC\textsubscript{refs}} $\uparrow$ & {SC\textsubscript{in}} $\uparrow$ & {R-1} $\uparrow$  & {R-2} $\uparrow$ & {{R-L}} $\uparrow$ & {QA} $\uparrow$ & {SC\textsubscript{refs}} $\uparrow$ & {SC\textsubscript{in}} $\uparrow$ \\
\hline \hline

\multirow{7}{*}{\rotatebox{90}{\textit{Extractive}}}  
& Random & 31.48 & 6.16* & 17.13* & 9.92* & 25.97* & 50.02* & 15.36 & 1.02* & 9.46* & 3.09* & 22.41* & 59.17* \\
   & Centroid & 31.54 & 7.68* & 17.79* & 14.17* & 25.82* & 53.99* & 19.00 & 2.00* & 11.21* & 3.89* & 23.47* & 64.63* \\
    & LexRank & 28.02 & 5.87* & 16.42* & 8.64* & 22.63* & 51.31* & 22.15 & 2.66* & 12.20\fs & 4.95\fs & 23.49* & 67.20* \\
    & QT & 38.21 & 10.28* & 21.50* & \textbf{17.12}\fs & 41.15\fs & \textbf{90.78}* & 21.31 & 1.51* & 11.41* & 3.62* & 22.42* & 66.21* \\
    & SemAE & 40.11 & 11.12\fs & 23.48\fs & 10.03* & 27.89* & 59.67* & 17.52 & 1.58* & 11.26* & 2.66* & 21.83* & 57.19* \\
    & \textsc{Hercules}\textsubscript{ext} & 41.71 & \textbf{13.15}\fs  & \textbf{24.43}\fs & 16.11\fs & \textbf{43.98}\fs & 84.27\fs & 23.22 & \textbf{3.04}\fs & \textbf{12.51}\fs & \textbf{6.94}\fs & \textbf{24.38}\fs & \textbf{84.05}\fs \\
   \hline
  \multirow{5}{*}{\rotatebox{90}{\textit{Abstractive}}} & CopyCat & 34.97 & 12.07* & 22.89* & 24.55\fs & 37.29* & 68.74* & 16.97 & 1.50* & 11.21* & 4.39* & 22.99* & 63.01* \\
  & InstructGPT & 38.15 & 9.05* & 22.35* &  16.06* & 25.97*  & 49.74* & 23.40  & 2.71* & 13.64* & 6.89\fs & 21.87* & 45.63* \\
  & BiMeanVAE  & 41.12 & 12.97* & 26.42\fs & 23.77\fs & 36.20* & 66.89*  & 21.42 & 2.04\fs & 12.49\fs & 5.66\fs & 21.78* & 52.54* \\
  & COOP & 41.04 & 13.53\fs & 26.56\fs & \textbf{25.21}\fs & 39.35* & 70.26* & \textbf{24.17} & \textbf{2.79}* & \textbf{14.12}* & 6.14\fs & 22.51* & 58.35* \\
  & \textsc{Hercules}\textsubscript{abs} & \textbf{45.48}\fs & \textbf{14.76}\fs & \textbf{27.22}\fs & 24.58\fs & \textbf{60.11}\fs & \textbf{92.04}\fs & 18.81 & 2.05\fs & 11.77\fs & \textbf{7.67}\fs & \textbf{25.23}\fs & \textbf{82.72}\fs \\

    \hline

   & (References) & - & -\fs & -\fs & 93.62\fs & 93.43\fs & 58.90\fs & - & -\fs & -\fs & 89.40\fs  & 86.68\fs & 65.59\fs  \\
   & (Oracle) &  & 45.02\fs & 53.29\fs & 31.74\fs & 69.59\fs & 64.14\fs & R1 & 14.36\fs & 26.04\fs & 14.04\fs & 26.38\fs & 76.35\fs \\

    \end{tabular}
    \caption{Results for automatic evaluation of summary generation. R-2 and R-L represent ROUGE-2/L F1 scores. QA indicates the F1 score of a question answering system attempting to answer questions generated from reference summaries, based on generated summaries. SC\textsubscript{refs} and SC\textsubscript{in} indicate degree of entailment (measured using SummaC) of generated summaries against reference summaries and input reviews respectively. Significant differences compared to each variant of \textsc{Hercules} according to a paired t-test $(p<0.05)$ are marked with an asterisk, and best results in each class are bolded. Overall, both variants of \textsc{Hercules} outperform comparison systems. In particular, summaries generated by \textsc{Hercules} score highest on SC\textsubscript{in}, indicating that they most strongly represent the information contained in the input reviews.}  
    \label{tab:automatic_general}
\end{table*}

\begin{table}[t!]
    \centering
    \small
    \begin{tabular}{l||@{~}r@{~~}r@{~}|@{~}r@{~~}r@{~~}r@{~}}
     &  \multicolumn{5}{c}{\textbf{\textsc{Space}\textsubscript{asp}}} \\
    \textbf{System} & {R-2} $\uparrow$ & {R-L} $\uparrow$ & {QA} $\uparrow$ & {SC\textsubscript{refs}} $\uparrow$ & {SC\textsubscript{in}} $\uparrow$ \\
\hline \hline

    QT\textsubscript{asp} &  10.24 & 22.64 & 16.28 & 33.05 & \textbf{77.32} \\
    \text{AceSum}\textsubscript{ext} & 12.10 &  27.15 & \textbf{20.15} & \textbf{38.04} & 67.48  \\
    \textsc{Hercules}\textsubscript{ext} & 7.93 & 19.96 & 13.84 & 26.12 & 66.64  \\
    \hline
    AceSum & \textbf{12.65} & \textbf{29.08} & 17.94 & {35.95} & 70.76   \\
    \textsc{Hercules}\textsubscript{abs} & {10.04} & 25.35 & 13.88 & 32.63 & {70.52} \\
    \hline
    (References) & - & - & 94.35 & 92.86 & 64.64   \\
    \end{tabular}
    \caption{ROUGE scores for controllable summarization, compared to the aspect-specific summaries in \textsc{Space}. Although not specifically designed for aspect-specific summarization, \textsc{Hercules} is nonetheless able to generate useful summaries about a specified aspect.}  
    \label{tab:aspect_scores}
\end{table}

\subsection{Automatic Metrics}

We use ROUGE F1 \cite[][\mbox{R-2/R-L} in Tables \ref{tab:automatic_general} and \ref{tab:aspect_scores}]{lin-2004-rouge} to compare generated summaries to the references, calculated using the `jackknifing' method for multiple references as implemented for the GEM benchmark \cite{gehrmann-etal-2021-gem}. To evaluate the faithfulness of the summaries, we use an automatic Question Answering (QA) pipeline inspired by \citet{fabbri-etal-2022-qafacteval} and \citet{deutsch-etal-2021-towards}: we use FlairNLP \cite{flair} to extract adjectival- and noun-phrases from the \textit{reference} summaries to use as candidate answers; we generate corresponding questions with a BART question generation model fine tuned on SQuAD \cite{lewis-etal-2020-bart, rajpurkar-etal-2016-squad}; finally, we attempt to answer these generated questions from the \textit{predicted} summaries, using a QA model based on ELECTRA \cite{clark2020electra,bartolo-etal-2021-improving}. We report the token F1 score of the QA model on the generated questions as `QA'.

We also evaluate the extent to which the generated summaries are entailed by both the reference summaries and the input reviews using SummaC \cite{laban-etal-2022-summac}, reported as SC\textsubscript{refs} and SC\textsubscript{in} respectively. SummaC segments input reviews into sentence units and aggregates NLI scores between pairs of sentences to measure the strength of entailment between the source reviews and generated summary. SC\textsubscript{in} is the only \textit{reference free} metric we use, and directly measures how well the generated summaries are supported by the input reviews. Since the references for AmaSum were constructed independently from the input reviews, we consider SC\textsubscript{in} to be our primary metric for AmaSum.

\subsection{Model Configuration}

We use a Transformer architecture \cite{Vaswani2017} for our encoder~$\phi(\textbf{z} | \textbf{x})$ and decoder~$p(\textbf{y} | \textbf{z})$. Token embeddings were initialized from BERT \cite{devlin-etal-2019-bert}\footnote{We experimented with using BERT as the encoder but found no significant improvement, since the discrete encoding is the main bottleneck in the model.}. We set the codebook size $K = 12$, with the number of levels $D = 12$, based on development set performance. Other hyperparameters are given in \Cref{app:replication}. Our code and dataset splits are available at \mbox{\url{https://github.com/tomhosking/hercules}}.
 
For \textsc{Space}, we generate summaries using 5 generic and 5 specific paths (\Cref{sec:summ_construction}). For AmaSum, which was constructed from a single verdict sentence followed by more specific pros and cons, we use 1 generic path and 13 specific paths.

\begin{table*}[ht]
\renewcommand{\arraystretch}{1.2}
    \centering
    \small
    \begin{tabular}{@{}m{2cm}@{~}|@{~}m{13.5cm}@{}}
    \textbf{System} & \textbf{Output} \\
    \hline\hline
\textit{Reference}  &  The staff were very friendly, spoke fluent English, and helped with our local transportation needs and restaurant recommendations. The entire hotel was very clean, and the rooms and bathrooms were cleaned every day. The room was of good size for Paris and included a balcony. The bathroom was good sized, fully equipped, and private. Breakfast was continental and perfectly adequate. The location is good.  \\ 
\hline
\textsc{Hercules}\textsubscript{ext}  &  The room was very small. The staff is very friendly and helpful. It is walking distance to the highlights of the Latin quarter but a few blocks away from the college crowd (a good thing). The rooms were clean. The breakfast was sparse in choices. The location was great, being close to the place Monge Metro station. Breakfast was served in the basement. The bathroom was clean. They spoke English. The cafe across the street was yummy.  \\ 
\hline
\textsc{Hercules}\textsubscript{abs}  &  The room was clean and comfortable. The staff was very friendly and helpful. Walking distance to everything. Breakfast was good. The hotel is in a great location, just a few minutes walk from the train station. Breakfast was fine. The room and bathroom were very clean. The staff spoke English and were very helpful. There is also a small restaurant on the ground floor.  \\ 


    \end{tabular}
    \caption{\textsc{Hercules} output summaries convey useful information without being overly specific or verbose.}
    \vspace*{-.2cm}
    \label{tab:output}
\end{table*}

\begin{table*}[ht]
\renewcommand{\arraystretch}{1.2}
    \centering
    \small
    \begin{tabular}{@{}m{2cm}@{~}|@{~}m{13.5cm}@{}}
    \textbf{Aspect} & \textbf{Output} \\
    \hline\hline
    \textit{Rooms} & The room was very small. We had a room facing the street. The room was dark and dingy. The room and bathroom were very clean. \\
    \hline
\textit{Food} & The coffee was undrinkable. The breakfast was a bit disappointing. There is also a small restaurant on the ground floor. Breakfast is served in the basement. \\
\hline
\textit{Location} & The hotel is in a great location, just a few minutes walk from the train station. The hotel is very basic. There is also a small restaurant on the ground floor. The location is very convenient. \\
    \end{tabular}
    \caption{Aspect-specific summaries from \textsc{Hercules}\textsubscript{abs} convey information specific to the desired topic.}
    \vspace*{-.2cm}
    \label{tab:aspectoutput}
\end{table*}

\section{Results}
\label{sec:results}

\paragraph{Automatic Evaluation}

The results in \Cref{tab:automatic_general} show that \textsc{Hercules} outperforms previous approaches on both datasets. On \textsc{Space}, \textsc{Hercules}\textsubscript{abs} achieves the highest ROUGE scores by some distance, and performs very well on all faithfulness metrics. 

On AmaSum, \textsc{Hercules}\textsubscript{ext} achieves higher ROUGE scores than \textsc{Hercules}\textsubscript{abs}; since the abstractive summaries are generated solely from the encodings, the decoder can sometimes mix up product types with similar descriptions (e.g., headphones and speakers) and is penalized accordingly. Since the references were not created from the input reviews, ROUGE scores are very low for all systems, and SC\textsubscript{in} is the most informative metric; both variants of \textsc{Hercules} achieve the highest scores. Surprisingly, a number of the systems achieve SC\textsubscript{in} scores higher than the references, indicating that they are generating summaries that are more grounded in the inputs than the gold standard. Systems that model the summary as a single sequence, like InstructGPT and COOP, achieve high ROUGE-L scores because they generate very fluent output, but are less informative and less grounded in the input reviews according to SC\textsubscript{in}, with InstructGPT scoring lowest on both datasets. \Cref{tab:output} shows an example of a summary generated by \textsc{Hercules} for an entity from \textsc{Space}. It covers a wide range of aspects, conveying useful information without being overly specific or verbose. We report additional examples in \Cref{app:examples}.

\begin{table}[t]
    \centering
    \small
    \begin{tabular}{lc||ccc}
 & \textbf{System} & \textbf{Info} $\uparrow$ & \textbf{Cohe} $\uparrow$ & \textbf{Conc} $\uparrow$  \\ 
\hline\hline 
 \multirow{5}{*}{\rotatebox{90}{\textit{ Extractive}}}  & Random & -9.68  & 0.20  & -3.21 \\ 
 & LexRank & -10.14  & -22.31  & -24.54 \\ 
 & QT & -8.05  & -5.44  & 0.51 \\ 
 & \textsc{Hercules}\textsubscript{ext} & \textbf{-0.10}  & \textbf{-1.99}  & \textbf{2.53} \\ 
 & (References) & 30.00  & 30.15  & 24.12 \\ 
\hline\hline 
 \multirow{5}{*}{\rotatebox{90}{\textit{ Abstractive}}}  & Random & -20.67  & -4.44  & -6.56 \\ 
 & InstructGPT & 0.22  & \textbf{3.78}  & \textbf{9.44} \\ 
 & COOP & -8.00  & -12.78  & -9.56 \\ 
 & \textsc{Hercules}\textsubscript{abs} & \textbf{1.44}  & -12.22  & -12.00 \\ 
 & (References) & 29.17  & 27.33  & 19.33 \\ 
\hline\hline 

    \end{tabular}
    
    \caption{Results of our human evaluation. Crowdworkers were asked for pairwise preferences between generated summaries in terms of their informativeness (Info), coherence \& fluency (Cohe) and conciseness \& non-redundancy (Conc). Higher scores are better, and best values within each system type are bolded (excluding references). Overall, \textsc{Hercules} generates more informative summaries than comparison systems.}
    \vspace{-.2cm}
    \label{tab:human_eval}
\end{table}

To evaluate the controllability of \textsc{Hercules}, we report the results of aspect-specific summarization on \textsc{Space} in \Cref{tab:aspect_scores} averaged across `rooms', `location', `cleanliness', `building', `service' and `food', with some example output shown in \Cref{tab:aspectoutput}. Despite not being specifically trained or designed to generate aspect-specific summaries, \textsc{Hercules}\textsubscript{abs} achieves reasonable scores across the range of metrics, and achieves comparable SC\textsubscript{in} scores to AceSum. \Cref{tab:aspectoutput} shows examples of aspect-specific summaries generated by \textsc{Hercules}\textsubscript{abs}, for the same entity. No sentiment-controlled reference summaries are available, but we show examples of sentiment-controlled output in \Cref{tab:sentiment}. We conclude that \textsc{Hercules} allows us to control the output of the model and generate summaries which focus on a specific aspect.


\paragraph{Human Evaluation}

Our goal is to generate summaries of hundreds of user reviews, but this makes human evaluation very difficult; it is not feasible to ask humans to keep track of the opinions expressed in hundreds of reviews. We are therefore limited to evaluation based on the references, but this is highly dependent on the reference quality. For AmaSum in particular the references are not grounded in the input reviews, and so the human evaluation is only indicative.

We recruited crowdworkers through Amazon Mechanical Turk, showed them a reference summary alongside two generated summaries, and solicited pairwise preferences along three dimensions: Informativeness, Conciseness \& Non-Redundancy, and Coherence \& Fluency. The full instructions are reproduced in \Cref{app:humaneval}. We gathered annotations for all 25~entities in the \textsc{Space} test set and 10~entities from each AmaSum domain, with 3~annotations for each. Extractive and abstractive systems were evaluated separately. The results in \Cref{tab:human_eval} show that both variants of \textsc{Hercules} produce summaries that are considered to be more informative than other systems, although this comes at the cost of slightly lower coherence.

\begin{table}[t]
    \centering
    \small
    \begin{tabular}{@{~}c@{~}||@{~}r@{~}r@{~}r@{~}|@{~}r@{~}r@{~}r@{~}}
    \textbf{Ablation} & \multicolumn{3}{c}{\textsc{\textbf{Space}}} & \multicolumn{3}{c}{\textsc{\textbf{AmaSum}}} \\
     & R-2 $\uparrow$ & R-L $\uparrow$ & SC\textsubscript{in} $\uparrow$ & R-2 $\uparrow$ & R-L $\uparrow$ & SC\textsubscript{in} $\uparrow$ \\
    \hline\hline
        \textsc{Hercules}\textsubscript{abs} & {14.76} & {27.22} & 92.04 & {2.05} & {11.77} & 82.72   \\
        \hline
        No norm loss & -1.32 & -1.02 & -1.86 & -0.01 & -0.11 & +1.08 \\
        No denoising & -1.99 & -2.85 & -5.34 & -0.17 & -0.23 & -7.75  \\
        Generic only & -0.82 & -0.59 & +3.28 & -0.66 & -0.70 & -14.77 \\
        Specific only & -1.49 & -3.41 & -11.24 & -1.15 & -2.18 & -9.91 \\
        VAE + k-means & -2.77 & -3.71 & -34.14 & -1.12 & -2.54 & -1.70  \\
    \end{tabular}
    \vspace{-0.1cm}
    \caption{Changes in key metrics for a range of ablations of the \textsc{Hercules}\textsubscript{abs} model. Removing the components tested leads to a drop in performance.}
    \vspace{-0.2cm}
    \label{tab:ablations}
\end{table}

\begin{table}[t]
    \centering
    \small
    \begin{tabular}{@{}m{1.2cm}@{~}|@{~}m{5.8cm}@{}}
    \hline\hline
\textit{Output} & Breakfast was good. \\
\hline
\multirow{6}{*}{\textit{Evidence}} & Breakfast was very good for us \\
& Breakfast offers a variety of things to eat. \\
& The buffet breakfast is varied and satisfying \\
& The buffet breakfast was all fresh food with a good choice \\
& Breakfast was good. \\
\hline\hline
\textit{Output} & Great camera for the price. \\
\hline
\multirow{5}{*}{\textit{Evidence}} & I like the camera. \\
& Overall a great camera at a good price. \\
& I like the range of the lens. \\
& Great camera. \\
& This is a good camera for the money. \\
\hline\hline
    \end{tabular}
    \vspace{-.1cm}
    \caption{Examples of evidence sets produced by \textsc{Hercules}. Each output sentence generated by the model is attached to a set of input sentences that share the same subpath.}
    \vspace{-.2cm}
    \label{tab:evidence_sets}
\end{table}

\paragraph{Attribution}

Since our approach is attributable and produces evidence sets alongside each abstractive summary sentence, we can evaluate the degree to which the generated sentences are supported by the evidence they cite. We used SummaC to measure the strength of entailment between each generated sentence and its evidence set, giving scores of 71.2 for \textsc{Space} and 46.8 for AmaSum. 
We also performed a human evaluation on a random sample of 150 output sentences, and found that generated sentences were supported by the majority of the associated evidence set 65\% of the time for \textsc{Space} and 57.3\% for AmaSum. We invite future work to facilitate this kind of evaluation and to improve on our level of factuality.

\begin{figure*}[ht!]
    \centering
    
    \includegraphics[width=0.99\textwidth]{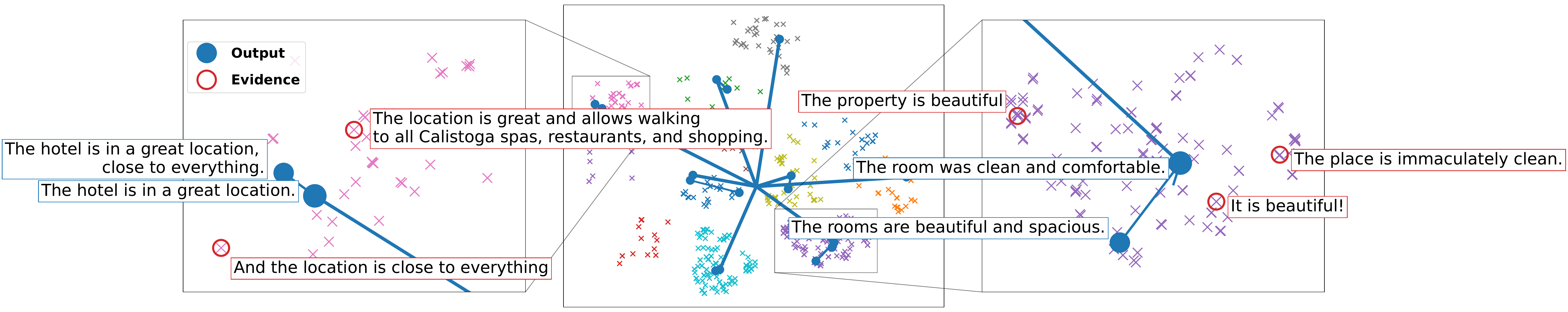}
    \caption{A t-SNE \cite{tsne} plot of the embeddings of all review sentences from a single entity from \textsc{Space}, where the colour of the points represents the top level code $q_1$. The summary subpaths are overlaid in blue, alongside output from different hierarchy depths. A selection of evidential inputs are circled in red.}
    \vspace{-0.4cm}
    \label{fig:tsne}
\end{figure*}

\paragraph{Ablations}

To evaluate to the contribution of each component towards the overall performance, we perform a range of ablation studies. \Cref{tab:ablations} shows the changes in key metrics for models trained without the norm loss and without the denoising objective. We also evaluate summaries generated using only the generic and specific subpath selection methods, rather than a combination of both. Finally, we evaluate the importance of learning the clusters at the same time as the model, rather than post-hoc: we train a model with the same training data and hyperparameters as \textsc{Hercules} but a \textit{continuous} encoding; use k-means clustering over sentence encodings to identify a set of centroids for each entity; and finally generate a summary by passing the centroids to the decoder. The results show that all components lead to improved summary quality. 
The centroids extracted from a continuous VAE using k-means may not necessarily correspond to a valid sentence, leading to poor quality output.

\begin{table}[t]
\centering
    \small
    \begin{tabular}{@{}m{2cm}@{~}|@{~}m{5.3cm}@{}}
    \hline\hline
    \textit{Input} & The staff was very helpful; the free breakfast was the best we had on this trip. \\
    \hline
    \textit{Output $(d=1)$} & Breakfast was good. \\
    \textit{Output $(d=2)$} & The continental breakfast was a joke. \\    
    \textit{Output $(d=3)$} & The breakfast was one of the best I have ever had. \\
    \textit{Output $(d=4)$} & The breakfast was one of the best I've had in a hotel. \\
\hline
    \multirow{7}{*}{\textit{Cluster $(d=3)$}} &
    Continental breakfast was the BEST so far on our trip!!! \\
& The staff was very helpful; the free breakfast was the best we had on this trip. \\
& The Cafe has the among the best breakfast and lunch in Vegas (closed for dinner). \\
\hline\hline
        \end{tabular}
        \vspace{-.1cm}
    \caption{An example of how our model encodes sentences at different granularities. As more levels are used, the output increasingly converges towards the meaning expressed by the input. We also show other input sentences that are assigned the same subpath (of depth = 3); despite very different phrasing, they convey a common opinion.}
    \vspace{-.2cm}
    \label{tab:analysis}
\end{table}

\paragraph{Analysis}

\Cref{tab:evidence_sets} shows examples of evidence sets, illustrating how \textsc{Hercules} is able to generate output that retains key information from the inputs, while discarding unnecessary detail. \Cref{tab:analysis} shows a breakdown of generated output at different granularities. Given the input sentence, we show the output of the decoder with subpaths of varying granularities, demonstrating how subpaths of increasing depth lead to more detailed output.

\Cref{fig:tsne} shows a t-SNE \cite{tsne} plot of the embeddings of all review sentences for a single entity from \textsc{Space}, with the summary subpaths overlaid on top in blue. We include a more detailed view of two summary subpaths (left and right panels), showing the increasing level of detail as more levels are specified. We also highlight sample input sentences from the evidence set (circled in red), demonstrating how the generated output can be attributed to input sentences conveying similar opinions.


\textsc{Hercules} is trained to reconstruct a target sentence from a source retrieved using tf-idf, but tf-idf is not sensitive to negation and does not distinguish between syntax and semantics. We observe that the model sometimes clusters sentences with superficially similar surface forms but different meanings. For example, ``The breakfast buffet was very good'' and ``The breakfast buffet was not very good either'' are assigned to the same path by our model.

The model is trained to generate output sentences based solely on the latent encoding: this is required to ensure that the model learns a useful encoding space. However, it also makes the model susceptible to some types of hallucination. Sentences about similar topics are likely to be assigned to the same paths, so the model may generate output that mentions a different entity of similar type (e.g., headphones instead of speakers).


\section{Related work}

Previous work has investigated aggregating user opinions in a latent space, but these approaches have generally been purely extractive for discrete spaces \cite{angelidis-etal-2021-extractive,basu-roy-chowdhury-etal-2022-unsupervised} or purely abstractive for continuous spaces \cite{iso-etal-2021-convex-aggregation}. Other approaches have either been supervised \cite{brazinskas-etal-2021-learning} or have selected `central' reviews to use as proxy summaries for training \cite{amplayo-etal-2021-aspect,amplayo2021unsupervised}, but they do not explicitly model the aggregation process. \citet{iso-etal-2022-comparative} propose a method for that highlights both common and contrastive opinions.

Hierarchical VQ was introduced by \citet{1171604} as `multistage VQ', with a set of codebooks fitted post-hoc to a set of encoding vectors, and further developed as `Residual VQ' by \citet{rvq} and \citet{10.1007/978-3-031-05933-9_17}. More recently, \citet{10.1109/TASLP.2021.3129994} and \citet{hosking-etal-2022-hierarchical} concurrently proposed methods for learning the codebook jointly with an encoder-decoder model. A form of hierarchical VQ 
has also been proposed in computer vision \cite{vqvae2}, but in their context the hierarchy refers to a stacked architecture rather than to the latent space. A separate line of work has looked at learning hierarchical latent spaces using hyperbolic geometry \cite{mathieu2019poincare,suris2021hyperfuture}, but the encodings are still continuous and not easily aggregated. 

The recent surge in performance of language models has led to a desire to evaluate whether the information they output is verifiable. \citet{DBLP:journals/corr/abs-2112-12870} propose a framework for post-hoc annotation of system output to evaluate attributability; we argue that it is better to have systems that justify their output as part of the generation process.

\section{Conclusion}

We propose \textsc{Hercules}, a method for aggregating user reviews into textual summaries by identifying frequent opinions in a discrete latent space. Compared to previous work, our approach generates summaries that are more informative, while also scaling to large numbers of input reviews and providing evidence to justify its output.

Future work could combine the improvements in attributability and scalability of our model with the fluency of systems that model summaries as a single sequence. Allowing the model to access the evidence sets during decoding could lead to improved output quality with less hallucination.

\section*{Limitations}

Since our approach identifies common opinions based on frequency of sentence encodings, we require a relatively large number of input sentences. We were not able to experiment with other popular datasets like Amazon \cite{10.1145/2872427.2883037}, Yelp \cite{meansum} or Rotten Tomatoes \cite{wang-ling-2016-neural} since these datasets only include a small number (usually 8) of input reviews.

The abstractive summaries are generated solely based on the latent encoding, and our model does not include a copy mechanism or attend to the original inputs when decoding. It therefore does not always generalize well to new domains. However, this limitation is mitigated by not requiring any labelled data during training: \textsc{Hercules} can easily be retrained on a new domain.

Generating output based only on latent encodings means that the model is also susceptible to hallucinating, since the output is less directly linked to the inputs. However, unlike other methods, \textsc{Hercules} provides evidence sets alongside the generated summaries, making it easier to check whether the output is faithful.

Finally, \textsc{Hercules} generates summary sentences independently, leading to summaries that are less coherent than approaches that model the summary as a single sequence. We welcome future work on combinining the relative strengths of each approach. We do not anticipate any significant risks resulting from this work.

\section*{Acknowledgements}
We thank our anonymous reviewers for their feedback. This work was supported in part by the UKRI Centre for Doctoral Training in Natural Language Processing, funded by the UKRI (grant EP/S022481/1) and the University of Edinburgh. Lapata acknowledges the
support of the UK Engineering and Physical Sciences Research Council (grant EP/W002876/1).

\bibliography{anthology,separator,main}
\bibliographystyle{acl_natbib}

\appendix

\section{Replication details}
\label{app:replication}

Models were trained on a single A100 GPU, with training taking roughly 24 hours for \textsc{Space} and 6 hours for each AmaSum domain.

The prompt used for InstructGPT evaluation was as follows:

\begin{quote}
    Review:
    
    [Review 1 text]
    
    Review:
    
    [Review 2 text]
    
    [...]
    
    Summarize these reviews:
\end{quote}


\begin{table}[ht]
    \centering
    \begin{tabular}{l|p{4cm}}
    \textbf{Parameter} & \textbf{Value} \\
    \hline
    Embedding dim. $D$ & 768 \\
    Encoder layers & 5 \\
    Decoder layers & 5 \\
    Feedforward dim. & 2048 \\
    Transformer heads & 8 \\
    Depth $D$& 12 \\
    Codebook size $K$& 12 \\
    Optimizer & Adam~\cite{adam} \\
    Learning rate & 5e-4 \\
    Batch size & 200 \\
    Token dropout & 0.2~\cite{tokendropout} \\
    Decoder & Beam search \\
    Beam width & 4 \\
    $\alpha_{init}$ & 0.5 \\
    $\tau_0$ & 1.0 \\
    $\tau_{min}$ & 0.5 \\
    $\gamma_{temp}$ & 33333 \\
    $\beta_{KL}$ & 0.0025 \\
    $\beta_{NL}$ & 0.05 \\
    $\gamma_{NL}$ & 1.5 \\
    \end{tabular}
    \caption{Hyperparameter values used for our experiments.}
    \label{tab:hyperparams}
\end{table}

\Cref{tab:hyperparams} show the hyperparameters used for our experiments. The Gumbel temperature was decayed from $\tau_0$ to $\tau_{min}$ according to 
\begin{multline}
    \tau = \max \big(\tau_0 \times \exp(-\frac{t}{\gamma_{temp}}), \tau_{min} \big ),
\end{multline}
in line with \citet{jang2016categorical}.

The model is sensitive to the initialization of the codebook; the initial embeddings should be located in roughly the same region of space as the output of the encoder, but should have sufficient variation so as to be informative for the decoder. Following \citet{9207145} we initialize the codebook on a unit hypersphere, to avoid the radial distance component dominating the angular component.

We used the default settings for SummaC \cite{laban-etal-2022-summac} as given on the project GitHub, using the SummaCConv variant trained on VitaminC \cite{schuster-etal-2021-get} and mean aggregation.

\section{Human Evaluation}
\label{app:humaneval}

The instructions given to crowdworkers were as follows:

\begin{quote}
    
In this task you will be presented with a number of summaries produced by different automatic systems based on user reviews. Your task is to select the best system summary based on the criteria listed below.

Please read the human summary first and try to get an overall idea of what opinions it expresses.

Please read the criteria descriptions and system summaries carefully, and whenever is necessary re-read the human summary.

Remember that you are being asked to rate the system, not the human summary.

\paragraph{Informativeness} Which system summary gives useful information that is consistent with the opinions in the human summary?

\paragraph{Conciseness \& Non-Redundancy}
Which system summary includes useful information in a concise manner and avoids repetitions?

\paragraph{Coherence \& Fluency}
Which system summary is easy to read and avoids contradictions?
\end{quote}

Crowdworkers were recruited from the UK and US, and were supplied with a Participant Information Sheet before being asked for their consent to participate. Crowdworkers were compensated \$0.30 per annotation which took approximately 1.5 minutes, corresponding to an hourly wage of \$12.00/hour. This exceeds the US federal minimum wage (\$7.25) at time of writing.

\section{Breakdown of Results}

We report the automatic evaluation scores broken down by AmaSum domains in \Cref{tab:amasum_rouge_by_cat} and \Cref{tab:amasum_qa_by_cat}, and the human evaluation results broken down by dataset in \Cref{tab:human_eval_full}.

\begin{table*}[t!]
    \centering
\small
    \begin{tabular}{cl||rr|rr|rr|rrHH}
& \textbf{{System}} &  \multicolumn{2}{c|}{\textbf{Electronics}} & \multicolumn{2}{c|}{\textbf{Home/Kitchen}} & \multicolumn{2}{c|}{\textbf{Shoes}} & \multicolumn{2}{c}{\textbf{Sports/Outdoors}} & \multicolumn{2}{H}{\textbf{Mean}} \\ 
 & &  R-2 $\uparrow$ & R-L $\uparrow$ & R-2 $\uparrow$ & R-L $\uparrow$ & R-2 $\uparrow$ & R-L $\uparrow$ & R-2 $\uparrow$ & R-L $\uparrow$ & R-2 $\uparrow$ & R-L $\uparrow$  \\ 
 \hline\hline
\multirow{6}{*}{\rotatebox{90}{\textit{Extractive}}} & Random &  0.95 & 9.51 & 1.09 & 9.78 & 0.76 & 8.41 & 1.29 & 10.17 & 1.02 & 9.46 \\ 
& Centroid &  1.78 & 11.53 & 2.50 & 12.47 & 1.63 & 9.43 & 2.07 & 11.41 & 2.00 & 11.21 \\ 
& LexRank &  2.47 & 12.18 & 3.22 & 12.84 & 1.96 & 10.51 & 3.00 & 13.29 & 2.66 & 12.20 \\ 
& QT &  1.55 & 10.95 & 1.79 & 12.15 & 1.23 & 11.13 & 1.46 & 11.43 & 1.51 & 11.41 \\ 
& SemAE &  1.32 & 10.97 & 2.37 & 12.95 & 1.32 & 9.64 & 1.32 & 11.48 & 1.58 & 11.26 \\ 
& \textsc{Hercules}\textsubscript{ext} &  3.29 & 12.48 & 3.19 & 12.89 & 2.67 & 11.75 & 3.00 & 12.93 & 3.04 & 12.51 \\ 
 \hline
\multirow{5}{*}{\rotatebox{90}{\textit{Abstractive}}}  & CopyCat &  1.46 & 11.92 & 2.11 & 11.86 & 0.98 & 9.00 & 1.46 & 12.07 & 1.50 & 11.21 \\ 
& InstructGPT &  2.83 & 13.89 & 2.99 & 14.39 & 2.23 & 12.52 & 2.80 & 13.76 & 2.71 & 13.64 \\ 
& BiMeanVAE &  2.32 & 12.42 & 2.32 & 12.93 & 1.46 & 11.80 & 2.05 & 12.81 & 2.04 & 12.49 \\ 
& COOP &  3.46 & 14.56 & 2.66 & 14.22 & 2.78 & 13.39 & 2.28 & 14.31 & 2.79 & 14.12 \\ 
& \textsc{Hercules}\textsubscript{abs} &  2.46 & 12.57 & 2.22 & 11.53 & 1.80 & 11.77 & 1.72 & 11.21 & 2.05 & 11.77 \\ 

    \end{tabular}

    \caption{Results for ROUGE scores with respect to references on AmaSum, broken down by product category.}
    \label{tab:amasum_rouge_by_cat}
\end{table*}

\begin{table*}[t!]
    \centering
\small
    \begin{tabular}{@{~}cl@{~}||@{~}r@{~}r@{~}r@{~}|@{~}r@{~}r@{~}r@{~}|@{~}r@{~}r@{~}r@{~}|@{~}r@{~}r@{~}r@{~}HHH}
& \textbf{{System}} &  \multicolumn{3}{@{~}c@{~}|@{~}}{\textbf{Electronics}} & \multicolumn{3}{@{~}c@{~}|@{~}}{\textbf{Home/Kitchen}} & \multicolumn{3}{@{~}c@{~}|@{~}}{\textbf{Shoes}} & \multicolumn{3}{@{~}c@{~}}{\textbf{Sports/Outdoors}} & \multicolumn{3}{H}{\textbf{Mean}} \\ 
 & &  {QA} $\uparrow$ & {SC\textsubscript{refs}} $\uparrow$ & {SC\textsubscript{in}} $\uparrow$ & {QA} $\uparrow$ & {SC\textsubscript{refs}} $\uparrow$ & {SC\textsubscript{in}} $\uparrow$ & {QA} $\uparrow$ & {SC\textsubscript{refs}} $\uparrow$ & {SC\textsubscript{in}} $\uparrow$ & {QA} $\uparrow$ & {SC\textsubscript{refs}} $\uparrow$ & {SC\textsubscript{in}} $\uparrow$ & {QA} $\uparrow$ & {SC\textsubscript{refs}} $\uparrow$ & {SC\textsubscript{in}} $\uparrow$  \\ 
 \hline\hline
\multirow{6}{*}{\rotatebox{90}{\textit{Extractive}}}   & Random &  1.90 & 22.55 & 57.27 &  1.97 & 22.87 & 57.87 &  2.80 & 22.20 & 62.80 &  5.68 & 22.03 & 58.72 & 3.01 & 22.41 & 59.17 \\ 
 & Centroid &  4.85 & 23.71 & 62.81 &  3.43 & 23.66 & 66.18 &  3.48 & 22.80 & 67.18 &  3.78 & 23.71 & 62.33 & 3.89 & 23.47 & 64.63 \\ 
 & LexRank &  5.12 & 24.04 & 68.36 &  5.09 & 22.49 & 57.91 &  5.09 & 25.22 & 84.44 &  4.49 & 22.19 & 58.11 & 4.96 & 23.49 & 67.20 \\ 
 & QT &  3.81 & 22.32 & 59.32 &  3.49 & 22.80 & 65.18 &  2.13 & 22.38 & 73.91 &  5.04 & 22.19 & 66.43 & 3.57 & 22.42 & 66.21 \\ 
 & SemAE &  0.41 & 22.07 & 55.64 &  4.55 & 21.81 & 53.13 &  5.05 & 21.82 & 63.59 &  0.61 & 21.61 & 56.39 & 2.72 & 21.83 & 57.19 \\ 
 & \textsc{Hercules}\textsubscript{ext} &  4.87 & 25.36 & 82.79 &  6.97 & 24.45 & 81.36 &  6.74 & 22.92 & 86.39 &  9.17 & 24.79 & 85.66 & 6.86 & 24.38 & 84.05 \\ 
 \hline
 \multirow{5}{*}{\rotatebox{90}{\textit{Abstractive}}} & CopyCat &  3.88 & 24.45 & 64.23 &  6.71 & 22.02 & 53.10 &  4.37 & 22.42 & 69.36 &  2.60 & 23.06 & 65.36 & 4.43 & 22.99 & 63.01 \\ 
 & InstructGPT &  5.61 & 22.42 & 47.50 &  3.80 & 21.55 & 44.20 &  9.54 & 22.10 & 47.57 &  8.63 & 21.40 & 43.24 & 6.87 & 21.87 & 45.63 \\ 
 & BiMeanVAE &  3.64 & 21.88 & 45.48 &  5.73 & 21.86 & 50.81 &  3.85 & 21.59 & 58.56 &  9.43 & 21.79 & 55.29 & 5.53 & 21.78 & 52.54 \\ 
 & COOP &  5.61 & 22.95 & 53.23 &  6.76 & 22.74 & 63.97 &  2.73 & 22.14 & 60.76 &  9.46 & 22.19 & 55.45 & 6.01 & 22.51 & 58.35 \\ 
 & \textsc{Hercules}\textsubscript{abs} &  6.54 & 25.55 & 79.49 &  8.98 & 25.25 & 82.15 &  5.97 & 24.59 & 85.09 &  9.19 & 25.53 & 84.16 & 7.60 & 25.23 & 82.72 \\ 
 \hline
 & (References) &  87.80 & 87.69 & 63.72 &  87.87 & 87.11 & 65.12 &  92.32 & 85.49 & 69.86 &  89.63 & 86.73 & 67.58 & 89.42 & 86.75 & 66.57 \\ 

    \end{tabular}

    \caption{Results for automatic faithfulness metrics on AmaSum, broken down by product category.}
    \label{tab:amasum_qa_by_cat}
\end{table*}

\begin{table*}[ht]
    \centering
    \small
    \begin{tabular}{lc||rrr|rrr||rrr}
        &  & \multicolumn{3}{c|}{\textsc{\textbf{Space}}}  & \multicolumn{3}{c||}{{\textbf{AmaSum}}}  & \multicolumn{3}{c}{\textbf{Overall}} \\ 
 & \textbf{System} & Info $\uparrow$ & Cohe $\uparrow$ & Conc $\uparrow$ & Info $\uparrow$ & Cohe $\uparrow$ & Conc $\uparrow$ & Info $\uparrow$ & Cohe $\uparrow$ & Conc $\uparrow$ \\

\hline\hline 
  \multirow{5}{*}{\rotatebox{90}{\textit{ Extractive}}}  & Random & -5.33 & -2.67 & -4.67 & -14.04 & 3.07 & -1.75 & -9.68  & 0.20  & -3.21 \\ 
 & LexRank & -27.33 & -39.33 & -44.67 & 7.05 & -5.29 & -4.41 & -10.14  & -22.31  & -24.54 \\ 
 & QT & -8.67 & -10.00 & -4.67 & -7.42 & -0.87 & 5.68 & -8.05  & -5.44  & 0.51 \\ 
 & \textsc{Hercules}\textsubscript{ext} & 1.33 & -2.00 & 0.00 & -1.54 & -1.98 & 5.05 & -0.10  & -1.99  & 2.53 \\ 
 & (Gold) & 40.00 & 54.00 & 54.00 & 20.00 & 6.30 & -5.75 & 30.00  & 30.15  & 24.12 \\ 
\hline\hline 
 \multirow{5}{*}{\rotatebox{90}{\textit{ Abstractive}}}  & Random & -18.67 & -2.67 & -11.33 & -22.67 & -6.22 & -1.78 & -20.67  & -4.44  & -6.56 \\ 
 & InstructGPT & -10.67 & 5.33 & 18.00 & 11.11 & 2.22 & 0.89 & 0.22  & 3.78  & 9.44 \\ 
 & COOP & -12.00 & -24.67 & -20.00 & -4.00 & -0.89 & 0.89 & -8.00  & -12.78  & -9.56 \\ 
 & \textsc{Hercules}\textsubscript{abs} & 4.67 & -16.00 & -18.67 & -1.78 & -8.44 & -5.33 & 1.44  & -12.22  & -12.00 \\ 
 & (References) & 36.67 & 38.00 & 32.00 & 21.67 & 16.67 & 6.67 & 29.17  & 27.33  & 19.33 \\ 
\hline\hline 

    \end{tabular}
    \label{tab:human_eval_full}
    \caption{Breakdown of human evaluation results by dataset.}
\end{table*}

\section{Example Output}
\label{app:examples}

\Cref{tab:sentiment} shows an example of generated summaries with sentiment control. We report additional examples of output summaries in \Cref{tab:additionaloutput}.

\begin{table*}[ht]
\renewcommand{\arraystretch}{1.2}
    \centering
    \small
    \begin{tabular}{@{}m{2.5cm}@{~}|@{~}m{12.5cm}@{}}
    \hline\hline
\textit{Rating = 1 (bad)} & \Tstrut Then it stopped working. It died in less than a year. Do not buy this machine. It didn't even last a year. Bought this in January 2017. \\
\hline
\textit{Rating = 5 (good)} & This is a great fan. It's very quiet. I love this fan. The light is bright. This is a very nice remote. \Bstrut \\
\hline\hline
  \textit{Rating = 1 (bad)} &  \Tstrut The carpet was stained and dirty. The room was filthy. The bathroom was disgusting. The staff was unfriendly and unhelpful. Avoid this hotel at all costs. \\
\hline
\textit{Rating = 5 (good) } & The hotel is very close to the airport. The shuttle service was great. The pool and hot tub were great. The food was delicious. The view from our room was breathtaking. \Bstrut \\
\hline\hline
    \end{tabular}
    \caption{Examples of sentiment-controlled summaries generated by \textsc{Hercules}, from \textsc{Space} and AmaSum.}
    \label{tab:sentiment}
\end{table*}

\begin{table*}[ht]
\renewcommand{\arraystretch}{1.2}
    \centering
    \small
    \begin{tabular}{@{}m{2cm}@{~}|@{~}m{13.5cm}@{}}
    \textbf{System} & \textbf{Output} \\
    \hline\hline

\textit{Reference}  &  \Tstrut The staff were helpful and friendly, especially in the spa! The hotel is consistently clean and well kept and The rooms are luxurious. The rooms have also been upgraded with new white comforters and new ruffle dusters. The hotel also features kitchens so that we can prepare breakfast and eat out in the beautiful garden and lunch at the picnic area so that we can barbeque, but the food in wine country is also wonderful! The location of the hotel is within walking distance to all the shops and restaurants. The grounds are beautiful with the heated indoor/outdoor pools .  \\ 
 \hline
\textsc{Hercules}\textsubscript{ext}  &  The staff was friendly and helpful. The pools well maintained. The rooms are clean and spacious. Thank you Roman Spa for being a great place to stay. It seems that the hotel is improving. The pools and hot tubs are fantastic. The grounds are very well maintained, with beautiful flowers, and waterfalls all throughout. Again, no service seems to be the mantra at the Roman Spa. The grounds were nice. The location is great and allows walking to all Calistoga spas, restaurants, and shopping.  \\ 
 \hline
\textsc{Hercules}\textsubscript{abs}  &  The staff was very friendly and helpful. the pool was great. The rooms were clean and comfortable. We had a great time. The hotel is in a great location. the pool and hot tub were great. The grounds are beautiful and well maintained. service was impeccable. The grounds are very nice and well maintained. The hotel is in a great location, close to everything. \Bstrut  \\

\hline\hline

\textit{Reference} & \Tstrut This uniquely shaped vacuum is ideal for sucking debris out of corners and tight spaces. V-shaped design drives debris in front of the vacuum for easier clean-up. Swivel head makes it easy to maneuver around furniture. Power cord extends 20 feet. Great for picking up debris in tight spots with no attachments needed. Runs quietly. Users say this vacuum is prone to clogging and requires routine maintenance \\
\hline
\textsc{Hercules}\textsubscript{ext} & Works great on my Handscraped, hardwood floors. This vacuum is not made for carpet, but works well on any hard flat surface. Lightweight and easy to maneuver. There are three areas that trap dirt and need to be cleaned in order to empty the vacuum. We bought this little vacuum for our new hardwood floors, and it's the best thing ever. My only complaint is the cord is pretty short. I love this vacuum!! This vacuum is amazing. Love this little vacuum. The suction is great! It picks up pet hair and dirt as advertised. And it picks up dust!! I have 4 cats and 2 dogs. \\
\hline
\textsc{Hercules}\textsubscript{abs} & The suction power is great. No scrubbing necessary. This little vacuum is amazing. Easy to maneuver. Great for hard floors. No more dust Bunnies! My dog sheds so much. This vacuum is amazing! Does not stay in place. Excellent customer service. The cord is too short. Love this vacuum! Great for pet hair. \Bstrut\\

\hline\hline 

\textit{Reference} & \Tstrut A gaming-specific external hard drive designed for the whole range of Xbox consoles. Two terabyte options (2TB and 4TB) offer plenty of space for installing games, apps, and files. Can be used with multiple Xbox consoles. The noise level of the hard drive is louder than most other options \\
\hline
\textsc{Hercules}\textsubscript{ext} & Very noticeably speeds up loading times for gaming on my Xbox. Tons of storage now for my Xbox one. No issues at all. All Seagate though. Used for Xbox one. It ’ s just been sitting on my TV stand connected to me Xbox one X. Plug and play. So we ended up buying an external drive. Super easy to install. I have about 50 games installed on this hard drive and still have 75\% of space left. Love this Ssd! Plenty of space for extra games. Stopped working after 2 years. This hard drive is awesome. \\
\hline
\textsc{Hercules}\textsubscript{abs} & Stopped working in less than a year. Easy to install and use. Works great with my Macbook pro. Plug and play. Plenty of room. This one does. Fast load times. Great for gaming. This hard drive is very fast. No SD card reader. This external hard drive is great. This Ssd is fast. Bought this for my bedroom. Great price and fast shipping. \Bstrut \\

\hline\hline

    \end{tabular}
    \caption{Additional examples of output from \textsc{Hercules} from both \textsc{Space} and AmaSum.}
    \label{tab:additionaloutput}
\end{table*}

\end{document}